\documentclass[letterpaper, 10 pt, conference]{ieeeconf}  
\usepackage{cite}    
\usepackage{color}
\usepackage{balance}
\usepackage{graphicx}
\usepackage[bookmarks=false]{hyperref}

\renewcommand{\theenumii}{\theenumi.\arabic{enumii}.}

\IEEEoverridecommandlockouts                              
\begin{document}

\title{\LARGE \bf DeepEvolution: A Search-Based Testing Approach \\ for Deep Neural Networks}

\author{Houssem Ben Braiek and Foutse Khomh \\
\emph{SWAT Lab., Polytechnique Montr\'{e}al, Montr\'{e}al, Canada}
    \\
     \emph{\{houssem.ben-braiek, foutse.khomh\}@polymtl.ca}
    }

\maketitle
\thispagestyle{empty}
\pagestyle{empty}

\begin{abstract}
The increasing inclusion of Deep Learning (DL) models in safety-critical systems such as autonomous vehicles have led to the development of multiple model-based DL testing techniques. One common denominator of these testing techniques is the automated generation of test cases, e.g., new inputs transformed from the original training data with the aim to optimize some test adequacy criteria.
So far, the effectiveness of these approaches has been hindered by their reliance on random fuzzing or transformations that do not always produce test cases with a good diversity.
To overcome these limitations, we propose, DeepEvolution, a novel search-based approach for testing DL models that relies on metaheuristics to ensure a maximum diversity in generated test cases. We assess the effectiveness of DeepEvolution in testing computer-vision DL models and found that it significantly increases the neuronal coverage of generated test cases. Moreover, using DeepEvolution, we could successfully find several corner-case behaviors. Finally, DeepEvolution outperformed Tensorfuzz (a coverage-guided fuzzing tool developed at Google Brain) in detecting latent defects introduced during the quantization of the models. These results suggest that search-based approaches can help build effective testing tools for DL systems.
\end{abstract}

\section{Introduction}
Deep Neural Networks (DNN)-based software systems are considered to be the next generation of software~\cite{soft2}, thanks to their innovative development paradigm, where the program logic is inferred automatically from data using statistical learning methods. 
Recently, they have been deployed in large-scale and critical systems such as self-driving cars. However, ensuring the quality assurance of DNN-based software is still very challenging as evidenced by recent deadly incidents with Uber's cars\footnote{\url{https://www.nytimes.com/2018/05/07/technology/uber-crash-autonomous-driveai.html}}.  
In fact, because of their non-deterministic nature and the absence of a reference oracle, it is very challenging to reason about the behavior of a DNN and hence to test it. Novel testing techniques are needed both during model engineering and deployment phases, to guarantee the reliability and robustness of in-production DNN-based software.
During the model engineering phase, developers need to assess the impact of their configuration choices carefully. The effectiveness of this assessment depends on the capability of testing data to trigger both the major functionalities of the model (regular cases) and the minor functionalities (corner cases). 
When deploying a trained DNN model in an embedded system or a cell phone, a quantization\cite{hubara2017quantized} of the model is often required to fit into this constrained environments (i.e., limited storage and computation resources). A post-deployment testing phase is required to assess the effect of this quantization on the reliability and the robustness of the model. In fact, the change in precision that occur during quantization increases the likelihood of coincidental correctness in the long sequences of linear and non-linear operations performed by DNNs. 
%
Therefore, the challenge is to generate testing inputs that are resilient to this phenomenon and, hence, capable of checking for the existence of inconsistencies and unexpected behaviors in the quantized model. 

In this paper, we propose DeepEvolution, a novel Search-based Software Testing (SBST) approach for DNNs models. DeepEvolution aims to detect inconsistencies and potential defects in DNN models. 
DeepEvolution relies on population-based metaheuristics to explore the search space of semantically-preserving metamorphic transformations. Using a coverage-based fitness function to guide the exploration process; it aims to ensure a maximum diversity in the generated test cases. 
We assessed the effectiveness of DeepEvolution on popular image recognition DNNs, i.e., LeNet\cite{lenet} and CifarNet\cite{cifar}. Results show that DeepEvolution succeeds in boosting the neuronal coverage of DNNs under test, finding multiple erroneous DNN behaviors. DeepEvolution also outperformed Tensorfuzz~\cite{tensorfuzz} in detecting latent defects introduced during quantization.\\ 
\textbf{The remainder of this paper is organised as follows.}\\ Section~\ref{sec:background} introduces the software testing concepts adapted by our approach. Section~\ref{sec:design} presents the testing flow of DeepEvolution. Section~\ref{sec:vision_instance} describes an instantiation of DeepEvolution to test computer-vision DNNs. Section~\ref{sec:evaluation} reports evaluation results, while Section~\ref{sec:threats} discusses threats to their validity. 
Section~\ref{sec:related_work} summarizes the most relevant related works and  Section~\ref{sec:conclusion} concludes the paper.

\section{Background On Software Testing}
\label{sec:background}
In the following, we briefly describe the fundamental software testing techniques that have been used and adapted by our proposed approach for DL testing. 
\subsection{Metamorphic Testing}
Metamorphic testing~\cite{metamorphic} is a pseudo-oracle testing technique that allows finding erroneous behaviors by detecting violations of identified metamorphic relations (MR). The first step of this technique is the construction of MRs that relate inputs in a way that the relationship between their corresponding outputs becomes known in prior, so the desired outputs for generated test cases can be expected. For example, a metamorphic relation for testing the implementation of the function $\sin(x)$ can be the transformation of input $x$ to $\pi-x$ that allows examining the results by checking if $\sin(x)$ differs from $\sin(\pi - x)$. Using MRs, a partial oracle can be generated automatically to test the program.   
Program's inputs are transformed following the MRs and expected outputs are computed. Any significant difference between the expected output and the output produced by the program under test indicates a defect in the program. 
\subsection{Code Coverage Criteria}
Test adequacy evaluation consists of assessing the fault-revealing ability of existing test cases. It is based on different adequacy criteria that estimate if the generated test cases are 'adequate' enough to terminate the testing process with confidence that the program under test is implemented properly. Code Coverage is one of the most used test adequacy evaluation criteria. It gauges the proportion of the program’s source code that is executed by test cases. In fact, test cases achieving high code coverage are more likely to uncover more defects, since they trigger more code execution paths. 
\subsection{Search-Based Testing}
The generation of test inputs with the aim of achieving high coverage is a hard problem that random testing often fails to solve (because of the size and complexity of software under test). Search-based software testing (SBST) techniques have been introduced to overcome the limitations of random testing. SBST techniques formulate the test coverage criteria as a fitness function that compares and contrasts candidate solutions in terms of their ability to cover new program's states. Using this fitness function, SBST techniques leverage metaheuristics, i.e., gradient-free optimizers requiring only few or no assumptions on the fitness function and inputs data, to drive the search into a promising area of the input space in order to generate effective test cases that help reaching an 
acceptable level of coverage in reasonable time.

\section{DeepEvolution: Testing Workflow}
\label{sec:design}
DeepEvolution aims to generate 
effective synthetic test cases from an existing test data. Instead of searching in the space of inputs of a model, DeepEvolution explores 
the space of transformations looking for 
interesting transformations that are able to provide effective test cases. Using a population-based metaheuristics, it iteratively evolves an initial set of candidate transformations, deriving new transformations that satisfy the following criteria: 
(i) they are significantly different from their parents to produce test data exhibiting new DNN's behaviors, and (ii) achieve high fitness values 
(to keep the search in relevant discovered regions). From one generation of candidates to another, DeepEvolution performs follow-up tests with the resulting transformed inputs and stores the failed tests that exhibit erroneous DNN’s behavior or induce a divergence between the original DNN and its quantized version. The evolution process stops when a predefined 
number of generations is attained. 
%
DeepEvolution requires as input: (i) a set of metamorphic transformations; 
(ii) a coverage-based fitness function capable of comparing different transformations based on the effectiveness of their produced test inputs; and (iii) a population-based metaheuristic. The fitness function should capture both local (neurons covered by a mutated input that were not covered by its corresponding original input) and global (neurons covered by a mutated input that were not covered by all previous test inputs) neuronal coverage. 
In the following we present an instantiation of DeepEvolution for computer vision.  
\section{DeepEvolution: Computer-vision Instance}
\label{sec:vision_instance}
Given the rapid progress and the impressive performance of DNNs in computer-vision tasks~\cite{computerVision}, we propose following the instantiation of DeepEvolution components for testing computer-vision DNN models. 
\subsection{Metamorphic Transformation}
First, we gather a list of parametric image-based transformations that can be organised in two groups: 
\begin{enumerate}
    \item \textbf{Pixel-value transformations: }change image contrast, image brightness, image blur, image sharpness and random perturbations within a limited interval.
    \item \textbf{Affine transformations:} image translation, image scaling, image shearing, and image rotation.
\end{enumerate}
Because each image-based transformation has a theoretical domain, which defines the interval of possible values of its parameters such as brightness factor or rotation angle $\theta$, when applying transformations, we need to take into account these domain boundaries to ensure that 
transformed inputs are semantically equivalent to the original ones. 
To infer the valid domain interval of each transformation's parameters, we manually tune them to set up the appropriate range of values, i.e., high and low boundaries, that preserves the input semantics before and after each transformation, with respect to the data distribution.

To enable a large-scale generation of synthetic inputs from existing labeled testing data, we build a compound metamorphic transformation that assembles all the image-based transformations described above, in order to enhance the changeability of mutations and the diversity of generated inputs. Its application on a given image consists of applying the supported pixel-value transformations in sequence, and then, performing each single affine transformation once, on the resulting mutated input. We opted for this conservative strategy that consists of applying only one affine transformation following the pixel-value transformations because applying multiple affine transformations at once would increase the chances of generating meaningless images, i.e., images that don't occur in real-world situations.\\ 
Our defined metamorphic transformation produces the following mutated inputs: inputs resulting from only pixel-value transformations and inputs that are the results of applying, separately, each one of the affine transformations.
To verify that generated inputs remain semantically equivalent to the original inputs, we compute a Structural Similarity Index (SSIM)~\cite{wang2004image} which assesses 
the similarity between two images based on the visual impact of three characteristics: luminance, contrast, and structure. We expect that pixel-based mutated inputs differs from their originals with respect to these characteristics, but a very low SSIM indicates that the new image looses mostly all the information inherited from its parent. Therefore, we reject mutated synthetic inputs for which SSIM values are lower than a tuned threshold. 
\subsection{DNN Coverage}
%
We adapt the Neuron Coverage (NC) metric proposed by Pei et al.~\cite{deepxplore} to capture two levels of coverage (i.e., local and global) for each test input. \\
\textbf{Local neurons coverage (NLNC):} this represents the new neurons covered by the mutated test input that have not been covered by its corresponding original test input.\\
\textbf{Global neurons coverage (NGNC):} this consists of the new neurons covered by the mutated test input that have not been covered by all the previous test inputs, including both genuine and synthetic test inputs.\\
We define the following fitness function: 
\begin{equation}
\vspace{-5pt}
\label{eq:fitness1}
  Fitness = \alpha \times NLNC + \beta \times NGNC
\vspace{-5pt}
\end{equation}
$\alpha$ and $\beta$ are weights assigned to each coverage measure. 
\subsection{Swarm-based Metaheuristics}
We encode our compound metamorphic transformations as a vector, where each component represents one parameter that may be related to either a pixel-value or an affine transformation. To ensure semantically preserving transformations, 
we use the valid domain intervals of transformations that we have already tuned manually to create the high and low boundaries vectors, defining the sub-space of exploration. 
We instantiate DeepEvolution using the following $7$ swarm-based metaheuristics. 
(1) Particle Swarm Optimization (PSO)\cite{PSO}, (2) Cuckoo Search Algorithm (CSA), (3) Bat Algorithm (BAT)\cite{BAT}, (4) Gray Wolf Optimizer (GWO)\cite{GWO}, (5) Moth Flame Optimizer (MFO)\cite{MFO}, (6) Whale Optimization Algorithm (WOA)\cite{WOA}, (7) Multi-Verse Optimizer (MVO)\cite{MVO}. 
These 
metaheuristics algorithms are flexible enough to be easily applicable to a broad class of constrained optimization problems involving high dimensional bounded real-valued vectors without any prior search space discretization. 
\section{Empirical Evaluation}
\label{sec:evaluation}
We assess the effectiveness of DeepEvolution through the following three research questions:\\
\textbf{RQ1: }How much can DeepEvolution increase the coverage of generated test cases? \\
\textbf{RQ2: }Can DeepEvolution detect diverse erroneous behaviors in DNN models?\\
\textbf{RQ3: }Can DeepEvolution detect divergences induced by DNN quantization?
\subsection{Experiment Setup}
\textbf{Datasets.} We selected the two popular publicly available datasets, MNIST\cite{mnist} and CIFAR-10\cite{cifar10}, as our evaluation subjects. Since neuronal coverage estimations and models post-execution analysis are computation-intensive tasks, we decided to take random samples from each of our studied test datasets as initial testing data. More specifically, we randomly selected two samples $D_1$ and $D_2$ from each dataset; with increasing size (i.e., respectively $50$ and $100$).\\
\textbf{DNNs.} For each dataset (i.e., MNIST and CIFAR-10), we took, respectively, the official open-source implementation of Tensorflow models, LeNet\cite{lenet} and CifarNet\cite{cifar} to allow the reproducibility of our results and comparisons with our approach.\\
\textbf{Settings of DeepEvolution.} We adopt the default configurations of metaheuristics (which is a conservative approach) and we choose $\alpha=0.1$ and $\beta=1.0$ for the fitness function, which is consistent with their corresponding measure magnitude and our priority of increasing the neuron coverage. We select the common hyper-parameters, $population size = 10$ and $max iterations = 10$. To avoid the effects of randomness, all results are averaged over $3$ runs or more.
\subsection{RQ1: DNN Coverage Increase}
\textbf{Motivation.} We aim to evaluate if the generated test data can help increasing the neuronal coverage, i.e., triggering neurons, which are not covered by the original test data.\\
\textbf{Findings.} \textbf{DeepEvolution significantly boosts the neuronal coverage.} Table \ref{DNN_coverage} shows the final neuronal coverage ratio achieved by each implemented swarm-based metaheuristic. The results show that the test data generated by all the studied metaheuristic algorithms significantly enhance the two coverage measures, as confirmed by the Wilcoxon Signed Rank tests. 
\begin{table}[h]
\vspace{-7pt}
\caption{The reached neuron coverage per metaheuristic}
\vspace{-7pt}
\centering
\begin{tabular}{c c c c c c c} 
\hline
\textbf{Meta} & 
\multicolumn{2}{c}{\textbf{MNIST}} & \multicolumn{2}{c}{\textbf{CIFAR-10}}\\
\cline{2-5}
\textbf{heuristic} & 
\textbf{$D_1$} &
\textbf{$D_2$} & 
\textbf{$D_1$} &
\textbf{$D_2$}\\
 \hline
 \textbf{Traditional}& 44.77& 50.89& 48.03& 53.16\\
 \hline
 \textbf{BAT}& 94.85& 96.35& 96.02& 97.99\\
 \hline
 \textbf{CS}& 94.74& 96.12& 96.78 &98.30\\
 \hline
 \textbf{GWO} & 92.77& 94.55& 96.32 &97.83\\
 \hline
 \textbf{MFO} & 93.11& 95.10& 95.64&97.52\\
 \hline
 \textbf{MVO} & 86.57& 90.06& 93.76&96.54\\
 \hline
 \textbf{PSO} & 91.11& 94.04& 95.50&97.49\\
 \hline
 \textbf{WOA} & 94.55& 95.91& 95.85& 97.68\\
 \hline
\end{tabular}
\vspace{-12pt}
\label{DNN_coverage}
\end{table}
Although the obtained neuronal coverage measures were generally high, the searching process reaches almost a stationary value when it could no longer improve the global coverage induced by the generated inputs, and as a consequence, its role becomes limited to only finding transformed inputs pushing the DNN to behave differently. Nevertheless, the augmentation of the original test data lend a refreshing boost that enabled the enhancement of neuronal coverage, which shows that adding more original instances enlarges the inputs search space to cover more possible test cases. This suggests that the quality of the initial input data is important 
for successfully covering the major patterns learned by the DNN model under test and increasing the chances of producing rare test inputs.

We noticed that \emph{MVO} performs slightly worse than the others. This can be explained by its tendency to exploit more around the best candidates previously found to converge quickly and not to go further in exploring solutions far from the best recognized regions. This characteristic is however helpful 
when a metaheuristic optimizer is used to find an optimal global solution at the end, but since our objective is to explore the maximum of relevant regions in the space, we need to increase the exploration ability of \emph{MVO}, we can fix its starting parameters that emphasize the exploration such as higher TDR (i.e., the distance of maximum variation around the best solution) and lower WEP (i.e., probability of generating new candidates around the best solution).

Furthermore, similarly to the usage of test coverage in traditional software testing, increasing the neuronal coverage has been shown to be an effective way to enhance the diversity of the generated inputs; allowing uncovering rare corner-case behaviors, and potentially, intensifying their defect-revealing ability. The effectiveness of our search-based approach in detecting defects is the main purpose of the next two research questions.
\subsection{RQ2: Detection of DNN Erroneous Behaviors}
\textbf{Motivation.} The objective is to assess the effectiveness of our approach in testing the robustness of the DNN; by finding misclassified synthetic inputs.\\
\textbf{Findings.} \textbf{DeepEvolution can effectively generate test cases that trigger erroneous behaviors of the DNN.} Table \ref{DNN_failed_tests} presents erroneous behaviors detected by each metaheuristic algorithm. 
\begin{table}[h]
\vspace{-7pt}
\caption{Number of erroneous behaviors per metaheuristic}
\vspace{-7pt}
\centering
\begin{tabular}{c c c c c c c} 
\hline
\textbf{Meta} & 
\multicolumn{2}{c}{\textbf{MNIST}} & \multicolumn{2}{c}{\textbf{CIFAR-10}}\\
\cline{2-5}
\textbf{heuristic} & 
\textbf{$D_1$} &
\textbf{$D_2$} & 
\textbf{$D_1$} & 
\textbf{$D_2$}\\
\hline
\textbf{BAT}& 488& 963& 317& 642\\
\hline
\textbf{CS} & 1567& 3499& 1533& 3001\\
\hline
\textbf{GWO} & 1298& 2411& 1046& 1929\\
\hline
\textbf{MFO} & 1343& 2955& 1098& 2387\\
\hline
\textbf{MVO} & 378& 774& 370& 764\\
\hline
\textbf{PSO} & 1116& 2913& 1108& 2403\\
\hline
\textbf{WOA} & 1702& 3601& 1122& 2335\\
\hline
\end{tabular}
\vspace{-12pt}
\label{DNN_failed_tests}
\end{table}
As all metaheuristic algorithms succeeds to reveal defects of the studied DNNs, it indicates that generating synthetic test inputs towards improving the neuronal coverage could trigger more states of a DNN, incurring higher chances of defect detection, which is consistent with the practical purpose of testing criteria used in traditional software testing. Indeed, the augmentation of sample data size, from $D_1$ to $D_2$ has significantly increased the number of erroneous behaviors detected. We obtained almost the double by doubling the input data size. 
This result suggests that DeepEvolution is capable of obtaining adversarial inputs for each original input and that the local coverage level integrated in the fitness function plays an important role in assessing how much the DNN's state of the transformed input is different from the state that resulted from the original input. Thus, it is capable of finding corner-cases testing inputs even if the global neuronal coverage reaches higher values; as evidenced by the increase in the triggered erroneous behaviors when augmenting the initial test data despite no significant improvement in the coverage between the two dataset samples. The results of \emph{MVO} reinforce the previous observation about their lack of exploration capability. 
The default implementation of \emph{BAT} also exhibits a similar insufficiency of diversification that could be remedied using adaptive rates of pulse emission $r$ (i.e., the probability of adjusting the found solutions) and loudness $A$ (i.e., the probability of generating a new candidate randomly). 
\subsection{RQ3: DNN Quantization Defects}
\textbf{Motivation.} The goal is to assess the usefulness of DeepEvolution in finding difference-inducing inputs that expose potential quantization defects.\\
TensorFuzz~\cite{tensorfuzz} performs a coverage-guided fuzzing process to generate mutated inputs that are able to expose disagreements between a DNN trained on MNIST (that is 32-bit floating point precision) and its quantized versions where all weights are truncated to 16-bit floating points. 
We use it as baseline to assess DeepEvolution. 
To ensure a fair comparison, we fix the configuration of TensorFuzz, including the data corpus size and number of mutations per element, in a way that the two approaches (TensorFuzz and DeepEvolution) produces the same number of test cases from each original test input.\\ 
\textbf{Findings.} \textbf{DeepEvolution can effectively detect defects introduced during DNN quantization, outperforming the coverage-guided fuzzing tool TensorFuzz.}
Table \ref{DNN_divergences} presents the number of synthetic test data that were able to induce a difference between the DNN's outcomes (difference-inducing inputs); exposing quantization defects.
\begin{table}[h]
\vspace{-7pt}
\caption{Number of quantization defects per metaheuristic}
\vspace{-7pt}
\centering 
\begin{tabular}{c c c c c c c c c} 
\hline
\textbf{Test Method} & 
\textbf{$D_1$} & 
\textbf{$D_2$}\\
\hline
 \textbf{TensorFuzz}& 8& 17\\
 \hline
 \textbf{BAT}& 32& 70\\
 \hline
 \textbf{CS} & 71& 136\\
 \hline
 \textbf{GWO} & 26& 103\\
 \hline
 \textbf{MFO} & 39& 78\\
 \hline
 \textbf{MVO} & 29& 66\\
 \hline
 \textbf{PSO} & 42& 86\\
 \hline
 \textbf{WOA} & 24& 69\\
 \hline
\end{tabular}
\vspace{-12pt}
\label{DNN_divergences}
\end{table}
As can be seen, all the implemented metaheuristics succeeded in exposing quantization defects and most of them outperformed TensorFuzz in terms of number of difference-inducing inputs found, confirming our intuition (and the motivation behind DeepEvolution) that by enabling the optimisation of coverage criteria, metaheuristic-based searching techniques can help increase diversity in generated test cases and hence improve their efficiency. 

\section{Threats to Validity}
\label{sec:threats}
In this section, we discuss potential threats to the validity of our work and highlight our mitigation measures. 

The selection of experimental subjects (i.e., dataset and DNN models) is a threat to the generalizability of our results. We mitigate this threat by using practical model sizes and commonly-studied MNIST and CIFAR-10 datasets. For each studied dataset, we choose to use official TF implementation with their corresponding configuration in order to avoid possible implementation bugs or misunderstanding issues that could hinder our evaluation process. Another threat could be the choice of parameters; we selected equal values for the two hyper-parameters, $population size = 10$ and $max iterations = 10$ because some metaheuristic algorithms rely on the iterative evolution of population and others rely on the availability of multiple candidates, so we choose the same value for them to make the evaluation as fair as possible. The selection of metaheuristic algorithms could be a threat to validity. We choose to implement swarm-based metaheuristics because of their randomness and non-deterministic nature that allowed them to be effective in resolving huge space problems. Furthermore, we implement several algorithms, because the No Free Lunch Theorem (NFL) states that no algorithm could outperform all other algorithms with regard to all optimization problems. 
However, when evaluating DeepEvolution with different metaheuristics, we do not compare their performance in solving our testing objective, since we use their default configuration and do not perform any hyperparameters tuning. In fact, we expect their performance to increase if they are tuned to fit our optimisation problem. 
The manual tuning of metamorphic transformations' domain and the threshold of SSIM could affect the validity of our results. 
To mitigate this threat, we selected a sample of our generated images using a confidence level of $95\%$ and an error margin of $5\%$, and verified them manually. We found them to be correct.


\section{Related Work}
\label{sec:related_work}
Pei et al.\cite{deepxplore} proposed DeepXplore, the first white-box testing framework specialized for DNN, which has two main components: (i) a new coverage metric specialized for DNNs (named neuron coverage) that estimates the amount of activated neurons and (ii) a differential testing component that uses multiple DNNs' implementations to solve the same problem (cross-referencing oracles in order to circumvent the lack of a reference oracle). 
Building on DeepXplore, Guo et al. proposed DLFuzz\cite{guo2018dlfuzz}, where mutation was restricted to the imperceptible pixel-value perturbations. 
%
%
Later, Tian et al. proposed DeepTest~\cite{deeptest}, a tool for automated testing of DNN-driven autonomous cars. In DeepTest, Tian et al. focus on generating realistic synthetic images by applying realistic image transformations like changing brightness, contrast, blurring, and fog effect to mimic different real-world phenomena. 
Based on the neuron coverage proposed by Pei et al., DeepTest performs a coverage-guided greedy search to finding realistic image transformations that can increase neuron coverage in a self-driving car DNNs. 
Odena and Goodfellow proposed a coverage-guided fuzzing framework named TensorFuzz~\cite{tensorfuzz}. This framework follows the same strategy of transforming new inputs from the original test data in way that maximize discovering novel DNN's states. 
However, it is based on a simple fuzzing process that consists in continuously adding random noises to inputs that previously triggered uncovered regions of the DNNs, with the hope of uncovering new states. We compared the performance of our proposed DeepEvolution to that of TensorFuzz an found it to be more effective at detecting
latent defects introduced during the quantization of DNNs.  

\section{Conclusion}
\label{sec:conclusion}
This paper presents DeepEvolution, a search-based DL testing approach that leverages semantically-preserving metamorphic transformations, DNN coverage criteria, and population-based metaheuristics. Our evaluation on computer-vision DNNs shows that DeepEvolution can improve the coverage of DNNs and successfully expose corner-cases behaviors. It also outperforms Tensorfuzz in detecting latent defects introduced during the quantization of models. 
\balance
\bibliography{main}{}
\bibliographystyle{IEEEtran}

\end{document}